\documentclass[10pt,twocolumn,letterpaper]{article}

\usepackage{cvpr}              

\makeatletter
\@namedef{ver@everyshi.sty}{}
\makeatother
\usepackage{tikz}

\usepackage{graphicx}
\usepackage{amsmath}
\usepackage{amssymb}
\usepackage{booktabs}
\usepackage{comment}
\usepackage{todonotes} 
\usepackage{multirow}
\usepackage[accsupp]{axessibility}

\usepackage[pagebackref,breaklinks,colorlinks]{hyperref}

\usepackage[capitalize]{cleveref}
\crefname{section}{Sec.}{Secs.}
\Crefname{section}{Section}{Sections}
\Crefname{table}{Table}{Tables}
\crefname{table}{Tab.}{Tabs.}

\def\cvprPaperID{12} 
\def\confName{CVPR}
\def\confYear{2023}

\begin{document}

\newcommand{\knowncl}{\mathcal{C}_\text{seen}}
\newcommand{\unknowncl}{\mathcal{C}_\text{unseen}}
\newcommand{\depthmap}{\mathbf{D}}
\newcommand{\rgbimage}{I} 
\newcommand{\cuecoord}{\mathtt{p}}
\newcommand{\segmask}{\mathbf{m}}

\title{Impact of Pseudo Depth on Open World Object Segmentation with Minimal User Guidance}

\author{Robin Schön \hspace{1.cm} Katja Ludwig \hspace{1.cm} Rainer Lienhart\\
Chair for Machine Learning and Computer Vision, University of Augsburg\\
{\tt\small \{robin.schoen, katja.ludwig, rainer.lienhart\}@uni-a.de}
}

\maketitle

\begin{abstract}
Pseudo depth maps are depth map predicitions which are used as ground truth during training. In this paper we leverage pseudo depth maps in order to segment objects of classes that have never been seen during training. This renders our object segmentation task an open world task. The pseudo depth maps are generated using pretrained networks, which have either been trained with the full intention to generalize to downstream tasks (LeRes and MiDaS), or which have been trained in an unsupervised fashion on video sequences (MonodepthV2). In order to tell our network which object to segment, we provide the network with a single click on the object's surface on the pseudo depth map of the image as input. We test our approach on two different scenarios: One without the RGB image and one where the RGB image is part of the input. Our results demonstrate a considerably better generalization performance from seen to unseen object types when depth is used. On the Semantic Boundaries Dataset we achieve an improvement from $61.57$ to $69.79$ \text{IoU} score on unseen classes, when only using half of the training classes during training and performing the segmentation on depth maps only.
\end{abstract}

\section{Introduction}
One of the central tasks of computer vision is the segmentation of images, and more particular, the segmentation of objects in these images. Most of the modern approaches to this task, however, require large amounts of annotated data. In the particular case of instance segmentation, each pixel of an object in the image has to be annotated in order to obtain suitable ground truth. Since this annotation process takes an inordinate amount of manual labeling time, methods have been crafted with the aim of at least partially alleviating this large quantity of work. These methods are known by the term of \emph{Interactive Segmentation} and try to enable the user to create a full segmentation mask of an object, while only requiring an input that is drastically faster to perform than manually annotating a full object segmentation mask. This input mostly consists of clicks, scribbles, bounding boxes or previously existing imperfect masks, which are, together with the image, fed into a neural network to predict a complete segmentation mask. 

\begin{figure}
    \centering
    {\setlength{\tabcolsep}{0.5pt}
    \begin{tabular}{ccc}
        \textbf{Pseudo Depth} & \textbf{Prediction} & \textbf{Ground Truth}\\
        \includegraphics[width=75pt, height=65pt]{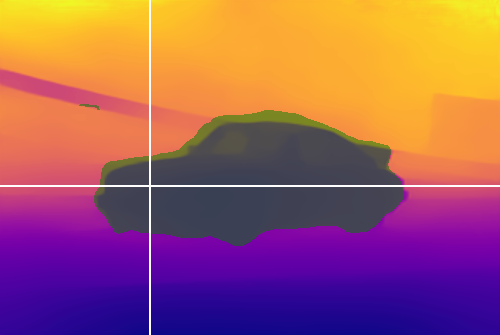} &  \includegraphics[width=75pt, height=65pt]{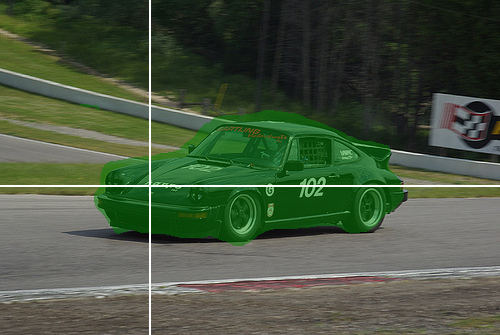} & \includegraphics[width=75pt, height=65pt]{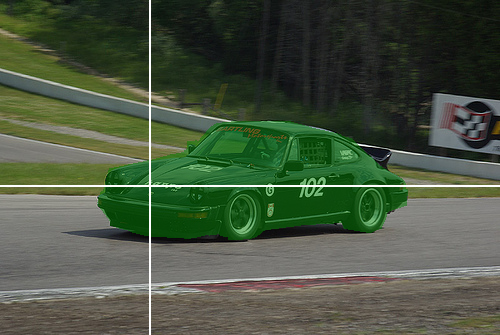} 
    \end{tabular}
    }
    \caption{Segmentation of a single object on a pseudo depth map. The object class is not in the training data, and the object to be segmented is indicated by a single coordinate (intersection point of the white crosshair). }
    \label{fig:intro_example}
\end{figure}

Generally, the task is independent of the object class. We only want to delimit the objects surface. In order to do that, we only need to recognize the objects shape and position, but not its class. Furthermore, Interactive Segmentation methods are meant to carry out predictions on new objects, which implies their usage on object classes that were not present during training. This observation invokes a need to distinguish between different object classes: 
\begin{itemize}
    \item $\knowncl$ are classes of objects for which we had labeled data during training.  
    \item $\unknowncl$ are classes of objects for which we did not have any labeled data during training. 
\end{itemize} 
Due to the unavailability of labeled data during training, we expect a lower performance on objects belonging to $\unknowncl$. 

In order to alleviate this problem we are going to use additional class independent information. In our particular case this information is depth. Due to depth being a strong cue regarding the shape of an object, depth provides helpful support to the segmentation task. We will measure the impact of depth usage in segmentation performance on the classes of $\knowncl$ in comparison to $\unknowncl$ in order to test the effectiveness of depth. We can, however, not expect depth ground truth to be available to use. This is a problem, that especially holds for classes that are entirely new. In order to tackle this problem, we will use pretrained monocular depth estimation networks to generate pseudo depth maps, so we will not be dependent on depth ground truth during training or testing. 

From the perspective of computational effort, we do not assume obtaining the depth maps to be an issue during usage, since the pseudo depth maps can always be generated much quicker than a user would be able to perform the input task. Alternatively, the depth can be precomputed once for the entire dataset before being shown to the user. This is especially useful if the user annotated the data on a low performance device. 

The nature of the issue investigated by us separates us from existing work on interactive segmentation. Most methods only compare different modes of interaction and user input, as well as how to process it. This however is not of interest to our experiments. We only want to explore the impact of depth. So in order to avoid skewing the experiments, we pick one particular simple type of user interaction and keep it for all experiments:  
The user merely indicates which object is to be segmented by single click on the surface of the object in the image. This also implies, that there is no repeated user interaction with the system, but only the initial indication of the object of interest.

In order to make sure, that the classes in $\mathcal{C}_\text{unseen}$ are completely new to the network, we train the networks from scratch. 
It  should be noted, that the aforementioned points would render a performance comparison with existing interactive segmentation methods superfluous.

We are going to show the effectiveness of such pseudo depth maps when segmenting objects that belong to known and unknown classes, by identifying them with a location on the object. In addition to that, we are going to show that high quality pseudo depth maps themselves suffice as an input for open world interactive segmentation. We will show that pseudo depth maps are not only a sufficient replacement for the classical RGB input, but do in fact outperform it in some cases. 

Our contributions can be briefly summarized as follows: 
\begin{itemize}
    \item We show that the drop in segmentation performance between known and unknown classes decreases if we grant our network access to depth information.
    \item We show that the addition of depth information, that has been acquired without any supervision, results in a performance increase on objects of unknown object classes. 
    \item We show that the complete replacement of RGB images by depth maps results in a performance increase for the segmentation of unknown single objects. 
    \item In some instances of our depth acquisition we use a newly collected set of indoor videos, where the camera moves considerably while the scene remains static. This property renders the videos useful for unsupervised depth estimation. We provide a link to the list of these videos.
    \url{https://github.com/Schorob/openhouse_videos}

\end{itemize}


\section{Related Work}
\subsection{Indicating Masks by Points}
Interactive segmentation aims at crafting methods which support a potential user in annotating segmentation masks. The most crucial element of such methods is a geometric cue indicating which object on the image is meant to be segmented. When it comes to the encoding of clicks we follow the practice proposed by the authors of \cite{reviving2021}, and use a distance transform, given to the network as an additional input channel, in order to encode the click. Since we are uniquely interested in the effect depth maps have on the segmentation performance, we refrain from comparing different input mechanisms. We try to keep our input method as simple as possible, and thus use a single click on the surface of the target object. It should be mentioned that some methods use negative clicks (\cite{Chen_2022_CVPR, reviving2021, hao2021edgeflow}) that would be used to exclude image parts. The existing literature is, however, not restricted to clicks. This can be seen in \cite{zhang2020interactive} for example, where the authors use a delimiting box that contains the desired object. 

Another distinction can be made between different types of interactive segmentation methods: Some make use of iterative user input (\cite{fbrs2020, reviving2021, li2018latent}), which means that the mask which has been guessed from the user input is repeatedly shown to the user. This enables the user to correct unsatisfying results. Other methods, however, aim at gaining a maximum of performance from the first user input by requiring clicks to be positioned at particularly useful locations (\cite{ucp_net, dextr}).  

Although not in the form of input, the authors of \cite{cheng2021pointly} show that single points on a surface of an object are useful cues to its segmentation mask by using points as a surrogate for complete mask labels.

\subsection{RGB-D Segmentation}
The general usage of depth information with the purpose of improving the segmentation of images has established itself as a common practice, also due to the increased availability of RGB-D semantic segmentation datasets (see \cite{sun3ddataset, nyudepthv2, b3dodataset, sunrgbd_dataset}). 
One common strategy is to use the available depth data as a form of additional input, in order to provide the network with additional information about the scene. The network then has to somehow fuse the input modalities for an improved segmentation output (\cite{10.1007/978-3-030-58621-8_33, hu2019acnet, liu2022cmx, wang2022tokenfusion, zhangattention, zhourgb}). 
In some works this fusion is realized by a form of self-attention (\cite{zhangattention, hu2019acnet, liu2022cmx, 10.1007/978-3-030-58621-8_33}), which considers the channel and the spatial dimension separately by re-weighting features along the given dimension.
This mechanism is similar to the feature reweighting that takes place in \cite{squeeze_and_excitation}. 
In other cases the attention mechanisms are based on transformer-like attention (\cite{liu2022cmx, wang2022tokenfusion}), which recombines the tokens in the feature tensor.

Depth maps have also been applied to the task of Salient Object Detection (\cite{fan2020bbsnet, zhang2020bianet}), where the pixels belonging to the most dominant (salient) object have to be determined.

Additionally, there has been a number of publications attempting to predict segmentation maps from only depth maps. The authors of \cite{hand_part_segmentation} use depth maps to segment parts of hands. In \cite{synthetic_garbage_segmentation}, the authors leverage synthetic data in order to train a segmentation network on garbage objects for the purpose of robot assisted waste disposal. 

To our knowledge, we are the first to predict user interaction guided segmentation masks on unknown objects, while only using depth instead of RGB as input modality.

\subsection{Monocular Depth Estimation}
Monocular depth estimation is a pixel-wise distance prediction task, from the camera's focal point. Additionally, the predictions take place on single images, which only allows for the estimation of relative depth values.

In order to be useful in downstream tasks, the MiDaS system (\cite{Ranftl2022, Ranftl2021}) has been trained on numerous datasets at the same time. 
Another depth estimator for zero-shot generalization is the LeRes (\cite{Wei2021CVPR}) system, which refines its predictions with a point cloud rectification system. 

While the aforementioned systems require the use of ground truth data during training, it is also possible to train a network for monocular depth estimation from video sequences alone. A considerable amount of works (\cite{Zhou_2017_CVPR, monodepth2, watson2021temporal, Gordon_2019_ICCV, Casser2019DepthPW, yin2018geonet, Zhou2019UnsupervisedHD, shu2020featdepth}) follow a similar strategy for training the network, warping temporally close images onto another. 
Specifically, in this paper we are going to use the MonodepthV2 framework (\cite{monodepth2}).

\subsection{Unseen Objects During Test Time}
The segmentation and localization of single objects can be carried out without any insight into which type of object is seen. The authors of \cite{kim2021oln} and \cite{Saito2021LearningTD} propose mechanisms for the purpose of object instance segmentation. In \cite{robot_stereo_imgs} additional geometric information for the segmentation of unknown objects is conveyed by the usage of stereo images as input. 
Most similar to our work, in \cite{synthetic_garbage_segmentation} the object that is to be segmented is determined by a single click. However, in contrast to our work, the authors are entirely concentrated on waste objects that are viewed from above.

\section{Method}

\subsection{Task Statement and Overview}
\label{subsec:overview}
\paragraph{Task Statement.} 
In the following we describe the task to which we refer to as Open World Interactive Segmentation. We aim to segment the object in an image which is identified by an arbitrary location lying on the object's surface. We assume that the object classes are divisible into two complementary sets: $\knowncl$ and $\unknowncl$. 
$\knowncl$ contains those classes of objects to which we will have access in the form of images with ground truth segmentation maps during training. More specifically, we have a training dataset 
\begin{equation}
    \mathcal{D}_\text{seen}^\text{train} = \left\{ (\mathbf{x}_i, \mathtt{p}_i, \mathbf{y}_i) \right\}_{i=1}^N
\end{equation}
where $\mathbf{x}_i \in \mathbb{R}^{H \times W \times 3}$ is an image that contains an object belonging to one of the known classes, $\mathtt{p}_i \in \{1, ..., H\} \times \{1, ..., W\}$ is a coordinate that is somewhere on the surface of said object, and $\mathbf{y}_i \in \{0, 1\}^{H \times W}$ is the ground truth segmentation mask of the object. Our system will be trained to predict $\mathbf{y}_i$ given $(\mathbf{x}_i, \mathtt{p}_i)$ as input, where $\mathtt{p}_i$ indicates which object on $\mathbf{x}_i$ is to be segmented. 

The classes in $\unknowncl$ contain objects for which we have no such training examples. However, since the object is indicated by a coordinate instead of a preset group of classes, our the segmentation task can also be performed on objects that have not been seen during training. During test time, this allows us to not only make predictions on input pairs 
\begin{equation}
    \mathcal{D}_\text{seen}^\text{test} = \left\{ (\mathbf{x}_i, \mathtt{p}_i) \right\}_{i=1}^{M_k}
\end{equation}
which correspond to object types seen during training, but also on inputs 
\begin{equation}
    \mathcal{D}_\text{unseen}^\text{test} = \left\{ (\mathbf{x}_i, \mathtt{p}_i) \right\}_{i=1}^{M_u}
\end{equation}
where the desired object on $\mathbf{x}_i$ is in $\unknowncl$. 

To test the effect of depth maps on this task we will compare the aforementioned scenario with two other scenarios: 
\begin{itemize}
    \item In the first scenario we assume to have additional access to depth maps $\mathbf{d}_i \in \mathbb{R}^{H \times W}$, wherein each pixel represents the scenes distance from the camera.  This leaves us with input triples $(\mathbf{x}_i, \mathbf{d}_i, \mathtt{p}_i)$. 
    \item In the second scenario, the depth maps $\mathbf{d}_i \in R^{H \times W}$ will \emph{replace} the RGB image completely, leaving us with inputs $(\mathbf{d}_i, \mathtt{p}_i)$ on which to carry out the segmentation.
\end{itemize} 

\begin{figure*}
    \centering
    {\setlength{\tabcolsep}{0.5pt}
    \includegraphics[width=0.85\linewidth]{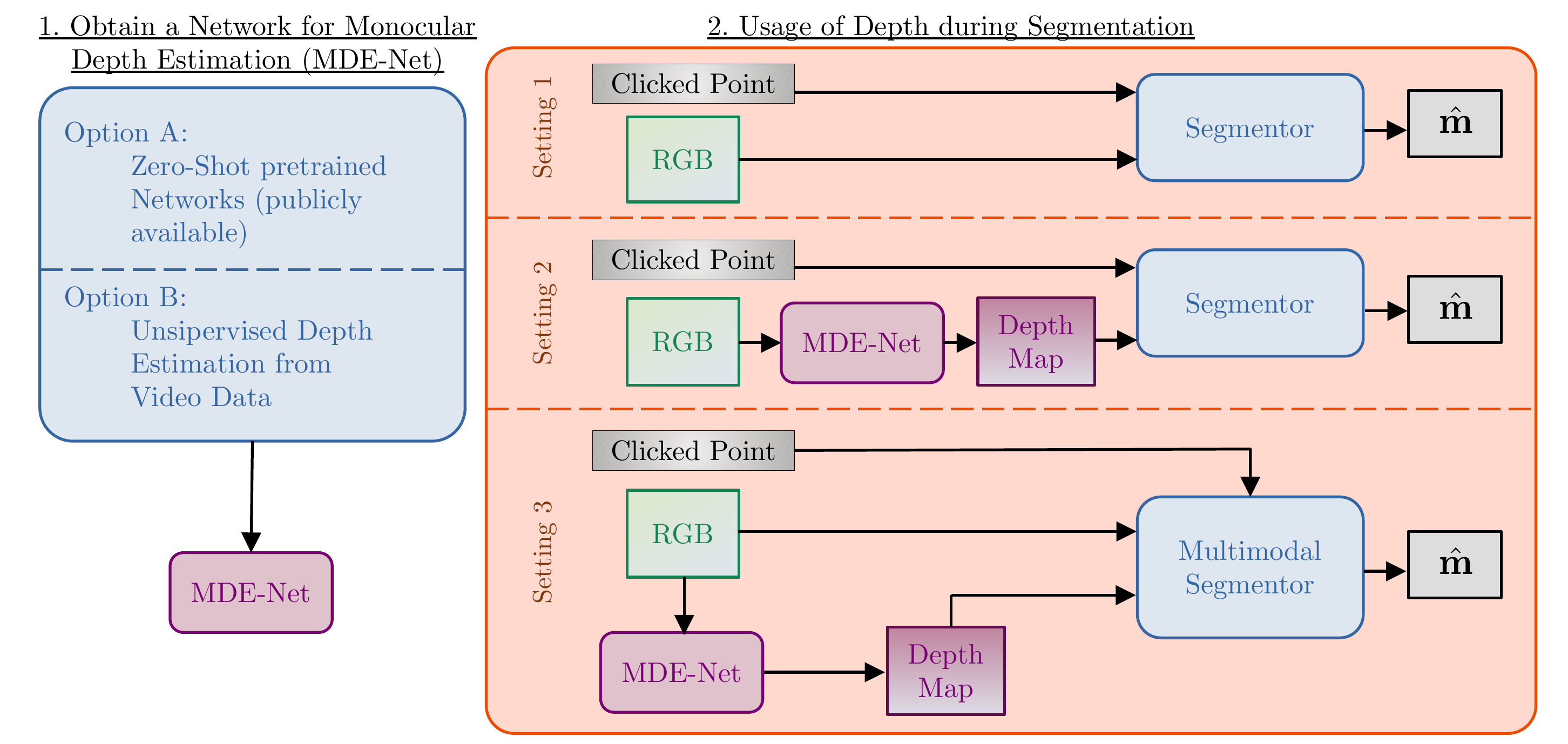}
    }
    \caption{This figure gives a general overview of our method. First, we obtain pretrained network for the task of monocular depth estimation (MDE-Net, left side). When it comes to the segmentation of objects we have three different settings. In the first setting, the network is given the RGB image and a clicked point, which indicates the object. In the second setting, we use the MDE-Net to generate pseudo depth maps, which then completely replace the RGB image. And in the third setting, we combine RGB and depth maps.}
    \label{fig:method_overview}
\end{figure*}

Figure \ref{fig:method_overview} provides an overview of our method.
Since our interaction mode is fixed to a single click, using the prevalent NoC@IoU metric (\eg \cite{reviving2021, fbrs2020}) would be inappropriate. Instead, we will measure the performance by the IoU after the single click has occurred.

\paragraph{Integration of Depth Information.}
When integrating depth information in our method we use two different strategies, depending on whether we want to provide the depth maps additionally or whether we want the depth maps to replace the RGB images entirely. 

In the former case we use the CMX model from \cite{liu2022cmx}, which uses a bifurcated backbone. This means that the two modalities, RGB images and depth maps, each have their own instance of the used backbone. After each stage in the network backbone, the feature maps are fused by a transformer inspired mechanism. The feature fusion is deliberately designed in such a way that both backbones have access to information originating from each other in the next stage. The fused features of each stage are given to the decoder after being subjected to two squeeze and excitation mechanisms \cite{squeeze_and_excitation}: The first one is channel-wise and the second one is over the spatial dimensions. 
In the latter case we use our standard image segmentation network architecture completely unaltered.

\paragraph{Training Loss.}
For the vast majority of training examples, the surface of the object that is to be segmented is considerably smaller than the background surface, which might incline our networks towards the background class. In order to avoid this problem we normalize our BCE loss. For an image of resolution $H \times W$, let $\segmask, \hat{\segmask} \in [0, 1]^{H \times W}$ be the ground truth and the predicted segmentation mask, respectively. Our loss is computed as  
\begin{equation}
\begin{split}
    \mathcal{L}_\text{balanced} =& - \sum_{i=1}^H \sum_{j=1}^W \frac{(1 - \segmask_{i,j}) \log(1 - \hat{\segmask}_{i, j})}{|\segmask = 0|}  \\
    +& \frac{\segmask_{i,j} \log(\hat{\segmask}_{i, j})}{|\segmask=1|}, 
\end{split}
\end{equation}
where $|\segmask=1|$ and $|\segmask=0|$ denote the number of foreground and background pixels. I.e. the background and foreground are equally weighted.

\subsection{Obtaining Depth without Ground-Truth}
\label{subsection:methoddepth}
Generally, we cannot assume to have access to ground truth depth information for arbitrary images. Those would either require additional hardware, such as a LiDAR sensor, or another image of the same scene. 
Therefore, for single image datasets, we make use of networks that have been trained for the task of monocular depth estimation. 

The information learned by the parameters of our depth estimation networks will be acquired in two different ways. 

\paragraph{Option 1: Zero-Shot Networks for Downstream Tasks.}
The first type of network is the result of supervised training on a large body of data in order to achieve zero-shot generalization on downstream tasks. These types of networks have been trained with multi-objective optimization goals on a combination of multiple datasets.
The LeRes system from \cite{Wei2021CVPR} is a combination of a classical monocular depth prediction network augmented with two additional point cloud networks. The latter two networks are used to refine the depth map predicted by the former one. The entirety of the system has been trained on five different datasets (\cite{3d_ken_burns, taskonomy2018, kim2018diml, hua2020holopix50k, Xian_2020_CVPR}). 
The MiDaS DPT network is based on a transformer like architecture (\cite{Ranftl2021, Ranftl2022}) and has been trained on ten different datasets (\cite{Xian_2018_CVPR, kim2018diml, Ranftl2021, MegaDepthLi18, wang2019web, tartanair2020iros, Xian_2020_CVPR, wang2019apolloscape, yao2020blendedmvs, wang2019irs}). We use the pretrained \texttt{DPT\_Large} version of this model, that is available via TorchHub.  

\paragraph{Option 2: Unsupervised Monocular Depth Estimation from Videos.}
The second possible option for obtaining pseudo depth maps consists in methods such as MonodepthV2 \cite{monodepth2}. This type of method leverages unlabeled video data, in order to train a neural network for the purpose of monocular depth estimation. A detailed description of the general mechanism providing this possibility is given in the Appendix 1. It should be noted, however, that despite requiring videos sequences during training, the resulting trained network will predict depth maps for single images.

\subsection{Marking of unknown objects}
The object to segment is identified by a single pixel location $\mathtt{p}$ on its surface.
In order to encode $\mathtt{p}$ in a way that is well interpretable by our network, we take inspiration from interactive segmentation literature (\cite{reviving2021}). We first compute the distance transform (\cite{felzenszwalb2012distance}) with respect to the point $\mathtt{p}$. This means we have a map $\mathcal{D}_\mathtt{p} \in \mathbb{R}^{H \times W}$ with the property 
\begin{equation}
    \forall \; (i, j)  \; : \; \mathcal{D}_\mathtt{p}(i, j) = \left\lVert \mathtt{p} - \begin{pmatrix} i \\j\end{pmatrix} \right\rVert_2. 
\end{equation}
It contains in each pixel the distance from our coordinate. This distance map is then concatenated as an additional channel to the image or depth map, respectively. 

\subsection{Training Details}
Unless stated otherwise, when using the SBD dataset, we always use the MiT-B0 backbone architecture which was published in \cite{xie2021segformer} in conjunction with the general SegFormer architecture. In cases where either only the RGB images or only the depth maps are used, we also use the SegFormer decoder, resulting in the standard architecture. However, when the RGB and depth information were both given by the network we also used the CMX mixing modules in between (see \cite{liu2022cmx}). For experiments that are carried out with the COCO dataset we use the MiT-B2 backbone due to its increased capacity. In all cases, our optimizer is the Adam optimizer (\cite{kimgma2015adam}) with learning rate $\alpha = 2 \cdot 10^{-4}$ and momentum parameters $\beta_1 = 0.9, \beta_2 = 0.999$. Our batch size is 8. In cases where the SBD dataset is used, we always train our model for exactly 300,000 iterations, and in cases where the COCO dataset is used we train our model for exactly 1,000,000 iterations. 

All our segmentation models are trained from scratch, to make sure that our unknown classes are \emph{really actually} unknown. We refrain completely from initializing our segmentation models with preexisting imagenet weights. The training data of the depth prediction networks on the other hand is very likely to have contained most, if not all, of the occurring object classes. This, however, does by no means contradict our hypothesis of depth maps being a suitable input modality for segmentation of unknown object classes. The segmentors will still be tasked with the segmentation of novel objects, based on the image's geometric data.


\section{Experiments}
\subsection{Datasets}
\paragraph{Segmentation Datasets.} The division of the segmentation task into seen classes $\knowncl$ and unseen classes $\unknowncl$ requires that the data we use is not only annotated with segmentation masks, but also with class labels. Thus, we make use of the instance segmentation datasets COCO (\cite{cocodataset}) and the Semantic Boundaries Dataset (SBD, \cite{BharathICCV2011}). 
When deciding which classes to train on, and which classes to test on, we always train on the classes for which there are the least annotated pixels, since less annotated pixels imply a smaller necessary annotation effort. For the SBD dataset \cite{BharathICCV2011} (which in total contains 20 classes) we use the 5, 7 and 10 least annotated classes, and for COCO the 5 least annotated classes. 

\paragraph{Origin of the Depth Maps.}  
Due to a lack of ground truth, all depth maps are predicted depth estimation networks.
Two of those models (MiDaS and LeRes) were pretrained in a supervised fashion and are discussed in further detail in Section \ref{subsection:methoddepth}. The Monodepth-based models had to be trained with video sequences. We utilize two different sources of data:
The first set of image sequences comes from the Mannequin Challenge Dataset \cite{mannequin_challenge}. The dataset is based on YouTube videos and provides high quality annotations, wherein not every frame is used, but instead just the frames which provide beneficial conditions for the task of depth estimation (moving camera + static scene). Additionally, the intrinsic camera parameters $K$ are annotated for every single frame. In total we train our depth estimation networks for 40 epochs on 106405 frames. 
We also wanted to test the effect of adding raw data. By this we mean the usage of the entire video (no labelling effort with respect to favorable or suitable time ranges) and unknown intrinsic camera parameters. For this purpose we collect a list of YouTube videos from the channel "OpenHouse24", which films the interior of empty houses. 
We use the entire videos, and extract every 3rd frame, since we consider adjacent frames too similar to allow for a meaningful optimization. The resulting sequence set consist of 259104 frames from 207 videos. We have published the link list to the videos, at \url{https://github.com/Schorob/openhouse_videos}. 
Due to the complete absence of humans in the dataset, we do not use this dataset alone, but only in combination with the Mannequin dataset. 
Confronted with the unavailability of usable camera parameters, we took the following substitute value
\begin{equation}
    K = \begin{pmatrix}
        1 & 0 & 0.5 \\
        0 & 1 & 0.5 \\ 
        0 & 0 & 1
    \end{pmatrix}
\end{equation}
as the intrinsic camera matrix.

\subsection{Segmentation in Depth Maps only}
The first results we are going to present are those concerning the case, where the RGB images are replaced completely by the pseudo depth maps of the corresponding image. Our metric is the \emph{Intersection over Union} between the predicted and the ground truth segmentation mask. The IoU is then averaged over all objects under consideration. 

We test our model on the validation split and train on the training set. For each of the two sets (train and val) we distinguish between the object types in $\knowncl$ and $\unknowncl$. We only train on those objects that are present in $\knowncl$, whereas we test on both unknown and known classes separately. We thus obtain two different IoU scores, $\text{IoU}_\text{seen}$ and $\text{IoU}_\text{unseen}$. 
Whenever we consider a certain number of classes as seen during training, the rest of the remaining classes are considered unseen. In the case of the 20 classes present in the SBD dataset, this results in (5 seen / 15 unseen), (7 seen / 13 unseen) and (10 seen / 10 unseen), respectively. In case of the COCO dataset, we only consider the 5 least annotated classes during training, leaving the remaining 75 classes unseen. 
In addition to the IoU score, which displays the pure performance, we are interested in the relative drop in performance 
\begin{equation}
    \Delta \% = \frac{\text{IoU}_\text{seen} - \text{IoU}_\text{unseen}}{\text{IoU}_\text{seen} } 
\end{equation}
on the unseen classes in comparison to the seen classes. This $\Delta \%$ value measures the effectiveness of depth when it comes to the generalization to unknown types of object. 
Table \ref{tab:SBD_depthonly} displays the results on the SBD dataset by the means of IoU and $\Delta \%$ scores. In the first line we have the experiments where only the RGB image has been used as an input, while in the remaining four lines only the depth has been used as an input.

\begin{table*}
    \centering
    \begin{tabular}{|l|c|c||c|c|c|c|c|c|c|c|c|}
    \hline
    \multicolumn{3}{|c||}{} & \multicolumn{3}{c|}{\textbf{5 Classes}} & \multicolumn{3}{c|}{\textbf{7 Classes}} & \multicolumn{3}{c|}{\textbf{10 Classes}} \\
    \hline
    Depth Origin & RGB & D & $\text{IoU}_\text{seen}$ & $\text{IoU}_\text{unseen}$ & $\Delta \%$ & $\text{IoU}_\text{seen}$ & $\text{IoU}_\text{unseen}$ & $\Delta \%$ & $\text{IoU}_\text{seen}$ & $\text{IoU}_\text{unseen}$ & $\Delta \%$ \\
    \hline 
    - & \checkmark &  & 62.84 & 56.71 & 9.75 & 65.16 & 59.66 & 8.44 & 67.84 & 61.57 & 9.24 \\
    \hline
    LeRes & & \checkmark & \textbf{68.36} & \textbf{66.3} & 3.01 & \textbf{71.51} & \textbf{69.23} & 3.19 & \textbf{74.1} & \textbf{69.79} & 5.82 \\
    MiDaS & & \checkmark & 64.89 & 62.89 & 3.08 & 66.62 & 65.1 & \textbf{2.28} & 69.95 & 66.77 & 4.55 \\
    \hline
    $\text{MD}_\textbf{mixed}$ & & \checkmark & 62.02 & 60.85 & 1.89 & 63.69 & 61.97 & 2.70 & 66.74 & 63.39 & 5.02 \\
    $\text{MD}_\textbf{clean}$ & & \checkmark & 62.65 & 62.22 & \textbf{0.69} & 64.78 & 63.23 & 2.39 & 65.53 & 63.99 & \textbf{2.35} \\
    \hline
    
    \end{tabular}
    \caption{This table displays the IoU scores for the segmentation task where the RGB input has been replaced with depth maps. The first line displays the results with RGB input only, while the the following lines display the results with \textbf{pseudo depth maps only}.  $\Delta\%$ denotes the performance drop between seen and unseen classes.}
    \label{tab:SBD_depthonly}
\end{table*}

In cases where the depth has been used, the column \emph{Depth Origin} indicates the way in which the pseudo depth maps have been obtained (see  Subsection \ref{subsection:methoddepth}). LeRes and MiDaS are trained with ground truth to generalize to downstream tasks. 
For $\text{MD}_\textbf{clean}$ and $\text{MD}_\textbf{mixed}$ we have trained a depth estimation model with the MonodepthV2 framework, which only required video sequences instead of ground truth. For $\text{MD}_\textbf{clean}$ we used the Mannequin dataset for which we had precisely annotated frame sequences and intrinsic camera parameters. In order to test the usability of depth obtained with raw, unfiltered sequences (complete videos from start to end; no availability of the intrinsic camera parameters), we have mixed the Mannequin dataset with the self-collected Openhouse dataset. The results for this configuration can be seen in the row with $\text{MD}_\textbf{mixed}$. 

\begin{table}
    \centering
    \begin{tabular}{|l|c|c||c|c|c|}
    \hline
    \multicolumn{3}{|c||}{}  & \multicolumn{3}{c|}{\textbf{5 Classes}} \\
    \hline
    Depth & \multirow{2}{*}{RGB} & \multirow{2}{*}{D} & \multirow{2}{*}{$\text{IoU}_\text{seen}$} & \multirow{2}{*}{$\text{IoU}_\text{unseen}$} & \multirow{2}{*}{$\Delta \%$}  \\
    Origin & & & & & \\
    \hline 
    - & \checkmark &  & \textbf{68.51} & 56.13 & 18.07 \\
    \hline 
    LeRes & & \checkmark & 63.92 & \textbf{58.39} & 8.65 \\
    MiDaS & & \checkmark & 59.51 & 56.27 & 5.44 \\
    \hline
    $\text{MD}_\textbf{mixed}$ & & \checkmark & 64.14 & 55.86 & 12.91 \\
    $\text{MD}_\textbf{clean}$ & & \checkmark & 60.24 & 55.08 & \textbf{8.57} \\
    \hline
    \end{tabular}
    \caption{This table displays the IoU scores where RGB images have been replaced with pure depth maps on the COCO dataset. The first line displays the results when only using the RGB images, while the other lines show results with \textbf{depth maps only} as input. $\Delta\%$ denotes the relative performance drop between seen and unseen classes. }
    \label{tab:COCO_depthonly}
\end{table}

\begin{table*}
    \centering
    \begin{tabular}{|l|c|c||c|c|c|c|c|c|c|c|c|}
    \hline
    \multicolumn{3}{|c||}{}& \multicolumn{3}{c|}{\textbf{5 Classes}} & \multicolumn{3}{c|}{\textbf{7 Classes}} & \multicolumn{3}{c|}{\textbf{10 Classes}} \\
    \hline
    Depth Origin & \multirow{2}{*}{RGB} & \multirow{2}{*}{D} & \multirow{2}{*}{seen} & \multirow{2}{*}{unseen} & \multirow{2}{*}{$\Delta \%$} & \multirow{2}{*}{seen} & \multirow{2}{*}{unseen} & \multirow{2}{*}{$\Delta \%$} & \multirow{2}{*}{seen} & \multirow{2}{*}{unseen} & \multirow{2}{*}{$\Delta \%$} \\
    / Config  & & & & & & & & & & & \\
    \hline 
    - / B0 & \checkmark &  & 62.84 & 56.71 & 9.75 & 65.16 & 59.66 & 8.44 & 67.84 & 61.57 & 9.24 \\
    - / B1 & \checkmark &  & 62.16 & 56.89 & 8.48 & 63.15 & 59.41 & 5.92 & 65.76 & 60.98 & 7.27 \\
    - / CMX & \checkmark &  & 61.05 & 56.65 & 7.21 & 66.75 & 60.13 & 9.92 & 67.61 & 61.42 & 9.16 \\
    \hline
    LeRes & \checkmark & \checkmark & \textbf{69.54} & \textbf{65.47} & 5.85 & \textbf{73.57} & \textbf{70.55} & 4.10 & \textbf{75.98} & \textbf{71.41} & 6.01 \\
    MiDaS & \checkmark & \checkmark & 69.42 & 65.39 & 5.81 & 71.24 & 68.91 & 3.27 & 74.19 & 69.55 & 6.25 \\ 
    \hline 
    $\text{MD}_\textbf{mixed}$ & \checkmark & \checkmark & 65.94 & 63.36 & 3.91 & 67.41 & 65.99 & \textbf{2.11} & 69.44 & 65.24 & 6.05 \\
    $\text{MD}_\textbf{clean}$ & \checkmark & \checkmark & 62.76 & 63.05 & \textbf{-0.46} & 68.13 & 66.05 & 3.05 & 69.23 & 66.42 & \textbf{4.06} \\
    \hline
    
    \end{tabular}
    \caption{This table displays the IoU scores for the segmentation task where RGB input has been augmented with depth. The first three lines show the results RGB input only, while the the following lines display the results for multimodal (RGB+Depth) models for different depth sources on the SBD dataset. The CMX model from \cite{liu2022cmx} was used to fuse the RGB images with the depth data. $\Delta \%$ denotes the relative performance drop between seen and unseen classes. }
    \label{tab:SBD_cmx}
\end{table*}

\begin{table}
    \centering
    \begin{tabular}{|l|c|c||c|c|c|}
    \hline
    \multicolumn{3}{|c||}{}  & \multicolumn{3}{c|}{\textbf{5 Classes}} \\
    \hline
    Depth & \multirow{3}{*}{RGB} & \multirow{3}{*}{D} & \multirow{3}{*}{$\text{IoU}_\text{seen}$} & \multirow{3}{*}{$\text{IoU}_\text{unseen}$} & \multirow{3}{*}{$\Delta \%$}  \\
    Origin & & & & & \\
    / Config  & & & & & \\
    \hline 
    - / CMX & \checkmark &  & \textbf{75.14} & 55.7 & 25.87 \\
    \hline 
    LeRes & \checkmark & \checkmark & 72.18 & \textbf{59.18} & 18.01 \\
    MiDaS & \checkmark & \checkmark & 71.36 & 56.72 & 20.52 \\
    \hline
    $\text{MD}_\textbf{mixed}$ & \checkmark & \checkmark & 69.25 & 57.25 & \textbf{17.33} \\
    $\text{MD}_\textbf{clean}$ & \checkmark & \checkmark & 72.01 & 58.05 & 19.39 \\
    \hline
    \end{tabular}
    \caption{This table displays the IoU scores, where RGB images and depth maps have both been used as input on the COCO dataset. In the first line "-/CMX" denotes the usage of RGB images as both inputs to the CMX based network. $\Delta\%$ denotes the relative performance drop between seen and unseen classes. }
    \label{tab:COCO_cmx}
\end{table}

\begin{figure*}
    \centering
    {\setlength{\tabcolsep}{0.5pt}
    \begin{tabular}{cccccc}
        \textbf{Depth} & \textbf{Prediction} & \textbf{Ground Truth} & \textbf{Depth} & \textbf{Prediction} & \textbf{Ground Truth} \\
        \includegraphics[width=80pt, height=60pt]{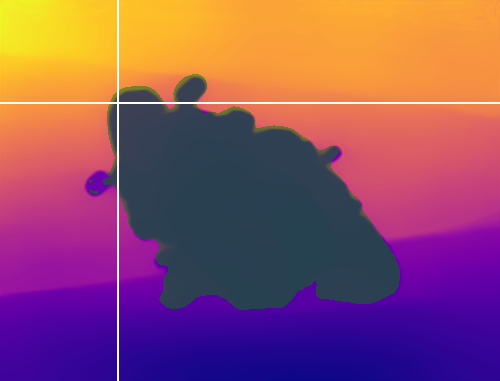} &  \includegraphics[width=80pt, height=60pt]{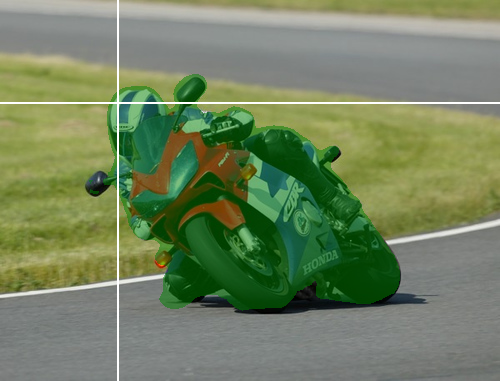} & \includegraphics[width=80pt, height=60pt]{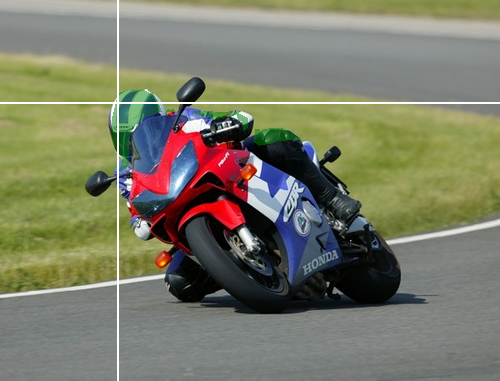} &
        \includegraphics[width=80pt, height=60pt]{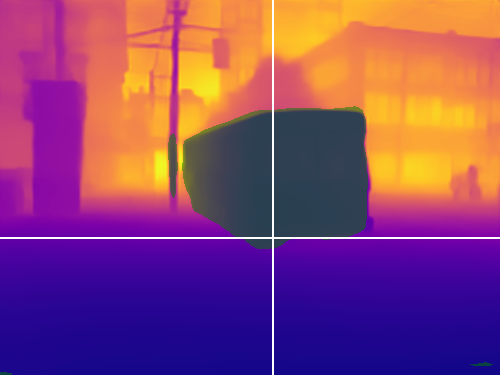} &  \includegraphics[width=80pt, height=60pt]{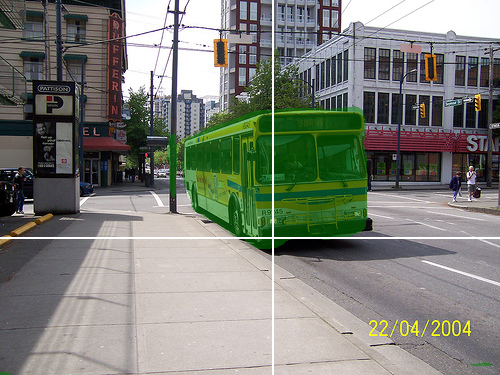} & \includegraphics[width=80pt, height=60pt]{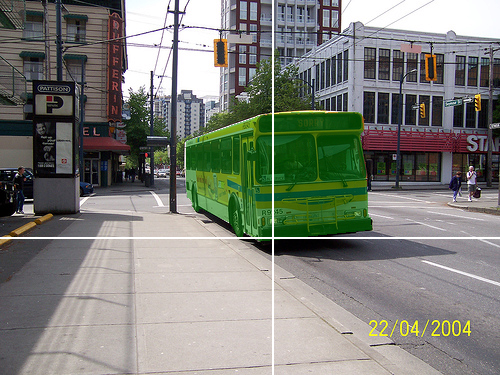} \\
        \includegraphics[width=80pt, height=60pt]{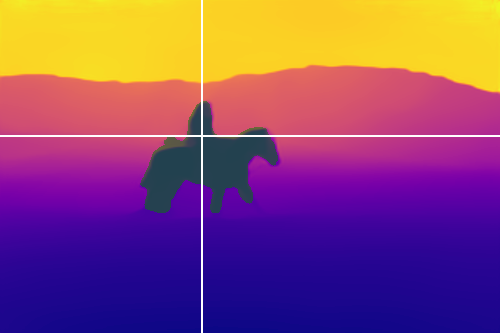} &  \includegraphics[width=80pt, height=60pt]{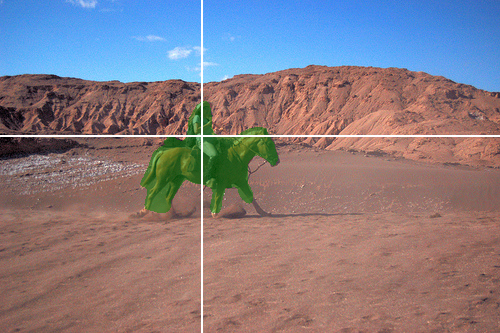} & \includegraphics[width=80pt, height=60pt]{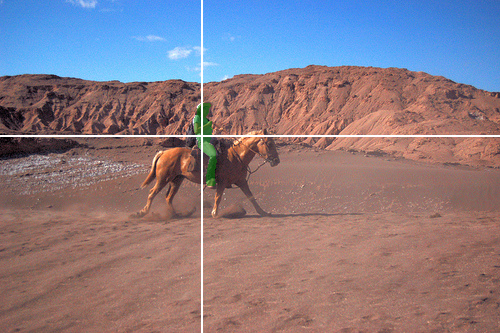} &
        \includegraphics[width=80pt, height=60pt]{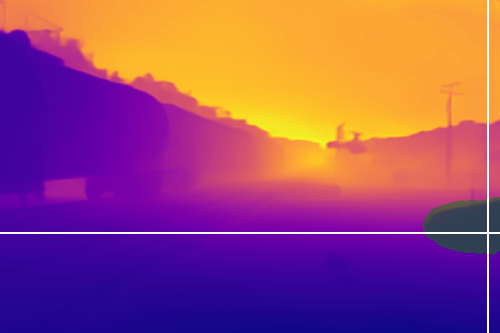} &  \includegraphics[width=80pt, height=60pt]{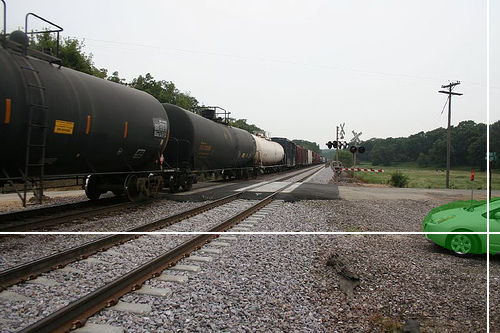} & \includegraphics[width=80pt, height=60pt]{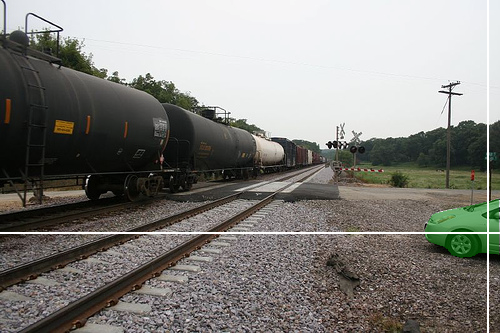} \\
        \includegraphics[width=80pt, height=60pt]{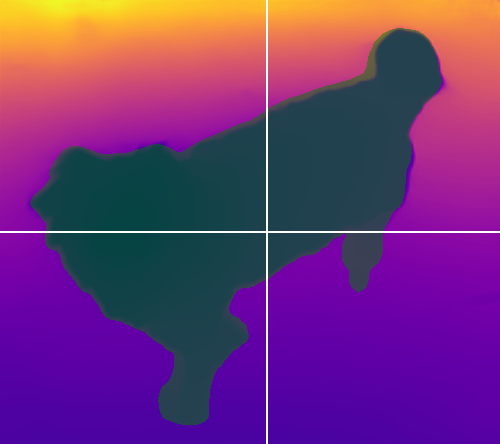} &  \includegraphics[width=80pt, height=60pt]{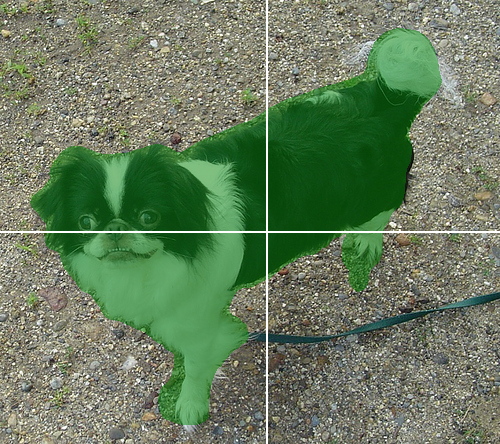} & \includegraphics[width=80pt, height=60pt]{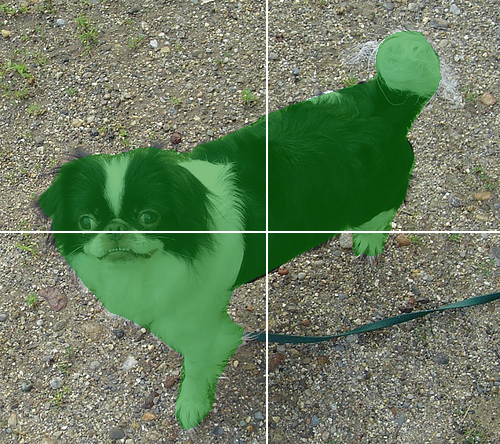} &
        \includegraphics[width=80pt, height=60pt]{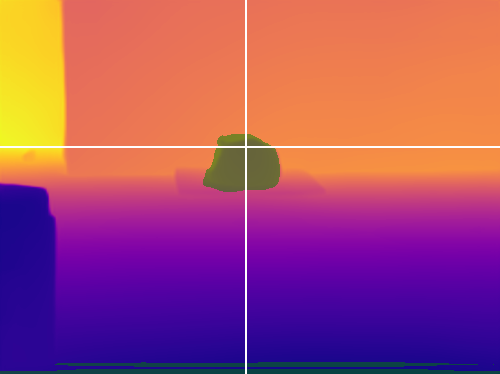} &  \includegraphics[width=80pt, height=60pt]{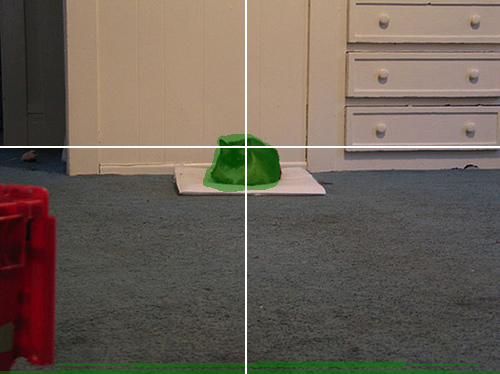} & \includegraphics[width=80pt, height=60pt]{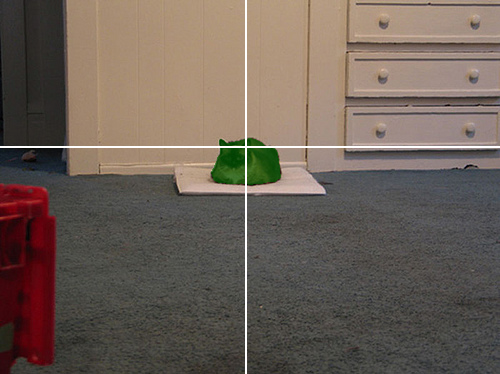} \\
    \end{tabular}
    }
    \caption{Qualitative examples (SBD dataset) for the case, where the segmentation prediction has been made entirely on the depth maps. All examples are of classes, which have been unseen during training. In the first and second row of the left side, we are able to see depth map induced failure cases. Especially the motor-cyclist has a similar depth to its vehicle, which incited the network to see them as a single cohesive surface. The white cross-hair indicates the cue coordinate $\mathtt{p}$.}
    \label{fig:qualitative_example}
\end{figure*}

Table \ref{tab:SBD_depthonly} shows the improvement of the performance on unseen classes in all cases. Especially in the case of 5 seen / 15 unseen classes, the IoU increases from 56.71 to 66.3 (by 16.91\%) when LeRes-based depth maps replace the RGB images. For 7 and 10 seen classes, the performance increases by 16.04\% and 13.35\%, respectively. Even when the depth estimation network has never seen any ground-truth label, as in $\text{MD}_\textbf{clean}$ and $\text{MD}_\textbf{mixed}$, we can observe an increase of the performance on the classes in $\unknowncl$.
This implies that depth constitutes a better modality for the segmentation of cohesive unseen objects than RGB itself, in case we already have a coordinate which gives us an anchor regarding the location of the object. 

In the cases of LeRes and MiDaS, which generate supervised depth maps, we can even see an improvement on classes which have already been seen in the ground truth data. 
These performance improvements allow for the conclusion that the geometric information of a single object constitutes a more useful information than the RGB image directly. The delimitation by the edges in the depth map give a very strong hint on the surface of the object in the image. In order to find a segmentation mask for an object whose location is already known, we only need to detect the border contours of the object. Information that only concerns the determination of the class itself is most likely unimportant. 

What can be evidently seen in almost all of the experiments is a deteriorating performance on the classes in $\unknowncl$ compared to those in $\knowncl$. The lower the relative performance drop $\Delta\%$, the better the generalization performance given the input. In Table \ref{tab:SBD_depthonly}, we can see that the $\Delta\%$ value is consistently lower in cases where the depth map has been used as input. This is probably the case, because depth maps are a decisively non-class specific type of information. Potential object textures and other non-geometric visual details, that would appear in RGB images (and which would not influence the segmentation surface, and thus are a risk factor of overfitting) are not present at all. 
Figure \ref{fig:qualitative_example} shows qualitative examples. The first two examples on the left of this figure show a potential downside of depth. As long as objects have a similar depth and are close to each other, they might be seen as a single object. 

On the COCO dataset we use the 5 least annotated classes as $\knowncl$ and the remaining 75 classes as $\unknowncl$. The results here can be seen in Table \ref{tab:COCO_depthonly}. In this case the RGB images seem to provide a better input to the segmentation taks on the seen classes $\knowncl$. We attribute this to the high variety of less salient objects in the COCO dataset. Additionally, since we use the least annotated 5 classes (hair dryer, toaster, baseball bat, sports ball, tooth brush) as seen classes, we happen to make use of rather obscure and small types of object during training. 
For the unseen classes, however, the depth maps seem to entail more useful information for the purpose of segmentation. The maximal improvement occurs when using the depth maps generated by the LeRes system from an IoU of 56.13 to 58.39. 
Also, in all cases of Table \ref{tab:COCO_depthonly}, the relative decrease $\Delta \%$ of the IoU between seen and unseen classes is smaller for the depth input than for the RGB images, affirming the better generalization capabilities exposed by depth maps.

\subsection{Combining RGB Images and Depth with Multimodal Segmentation Networks}

The second type of depth utilization consists in combining depth maps and RGB images by using the CMX architecture as described in Subsection \ref{subsec:overview}. We first look at the results for the SBD dataset (see Table \ref{tab:SBD_cmx}). 
Due to the CMX architecture having more parameters than the standard SegFormer-B0 (12.1 mio. vs. 3.7 mio.), we do not only compare these architectures. We also compare CMX to the SegFormer-B1 (13.7 mio. parameters) and the same CMX architecture, which gets two copies of the RGB input, to assure that we are dealing with comparably powerful networks. The configurations for the different RGB models are respectively called "B0", "B1" and "CMX" in Table \ref{tab:SBD_cmx}. 
As we can see by the results, the increased number of parameters in the RGB-only configurations does not result in a considerably better performance without the additional depth information. 

Providing depth maps and RGB images, however, leads to an increased IoU score. Also, the generalization performance between the seen and unseen classes improves, as can be seen by the smaller relative drops $\Delta \%$ . The strongest reduction of the generalization gap can be observed when using depth maps originating from the $\text{MD}_\textbf{clean}$ model, where the depth-based model peforms even better on unseen classes compared to seen classes (from $\Delta\% = 9.75$ to $\Delta\% = -0.46$). This can be attributed to the larger surfaces of the unseen classes, which makes depth based segmentation tendencially easier.

The results using the CMX architecture on the COCO dataset (see Table \ref{tab:COCO_cmx}) point in the same direction, as the ones obtained using only the depth maps as input. The usage of RGB inputs only is represented as a CMX network which receives a copy of the image for each of the backbones (see -/CMX in Table \ref{tab:COCO_cmx}). This configuration performs best (IoU of 75.14) on the classes, that were seen during training. The depth augmented network configurations, however, yield constantly better performance on the unseen classes. The maximum improvement occurs when using the LeRes based pseudo depth maps from 55.7 to 59.18 IoU. Furthermore, the depth helps diminishing the generalization gap in all cases it was used. The strongest drop in the generalization gap between seen and unseen classes can be observed using the depth maps from $\text{MD}_\textbf{mixed}$ from $\Delta \% = 25.87$ to $\Delta \% = 17.33$.


\section{Conclusion}
In this paper we have investigated and analyzed the usefulness of depth for the purpose of segmenting types of objects, that were never seen during training. 
A particular characteristic of our work is, that we assume no access to ground truth depth maps. 
The depth estimations were exploited in two different ways: 1) Depth replaced the RGB images in order to segment the indicated object, rendering it the only input modality to the network. 2) Depth was - in addition to the RGB images - given to a multi-modal fusion network with two backbones. 
We have shown that the segmentation of novel objects on depth maps only is not just viable, but considerably improves the generalization abilities from seen to unseen classes. On the SBD dataset we even obtained results, where depth maps performed better as input modality in comparison to RGB images.

{\small
\bibliographystyle{ieee_fullname}
\bibliography{egbib}
}

\twocolumn[{
	\makeatletter
	\begin{center}
		{\Large \bf Impact of Pseudo Depth on Open World Object Segmentation with Minimal User Guidance \\\large Supplementary Material\par}
		\vspace*{24pt}
		
		\large
		\lineskip .5em
		\begin{tabular}[t]{c}
			Robin Schön \hspace{2cm} Katja Ludwig \hspace{2cm} Rainer Lienhart\\
			Machine Learning and Computer Vision Lab, University of Augsburg\\
			{\tt\small \{robin.schoen, katja.ludwig, rainer.lienhart\}@uni-a.de}
		\end{tabular}
		\par
		
		\vskip .5em
		\vspace*{12pt}
	\end{center}
	\makeatother}]
\setcounter{figure}{0} 
\renewcommand{\thefigure}{S.\arabic{figure}}

This section aims at acquainting the reader with the general idea of unsupervised monocular depth estimation. This task consists in the utilization of unannotated video data with the purpose of training a network for the task of monocular depth estimation. It should be explicitly mentioned, that despite the necessity of video data during training, the resulting depth estimation network will be trained to predict depth maps for single images. This renders the depth estimator useful for downstream tasks on images, which is especially useful for our purposes. 
Although we mention the existence of a considerable amount of literature (see \cite{zhou2017unsupervised, godard2019digging, watson2021temporal, Gordon_2019_ICCV, Casser2019DepthPW, yin2018geonet, Zhou2019UnsupervisedHD, shu2020featdepth}), we will specifically use the MonodepthV2 framework as described in \cite{godard2019digging}. 

\begin{figure}
\centering
\includegraphics[width=0.9\linewidth]{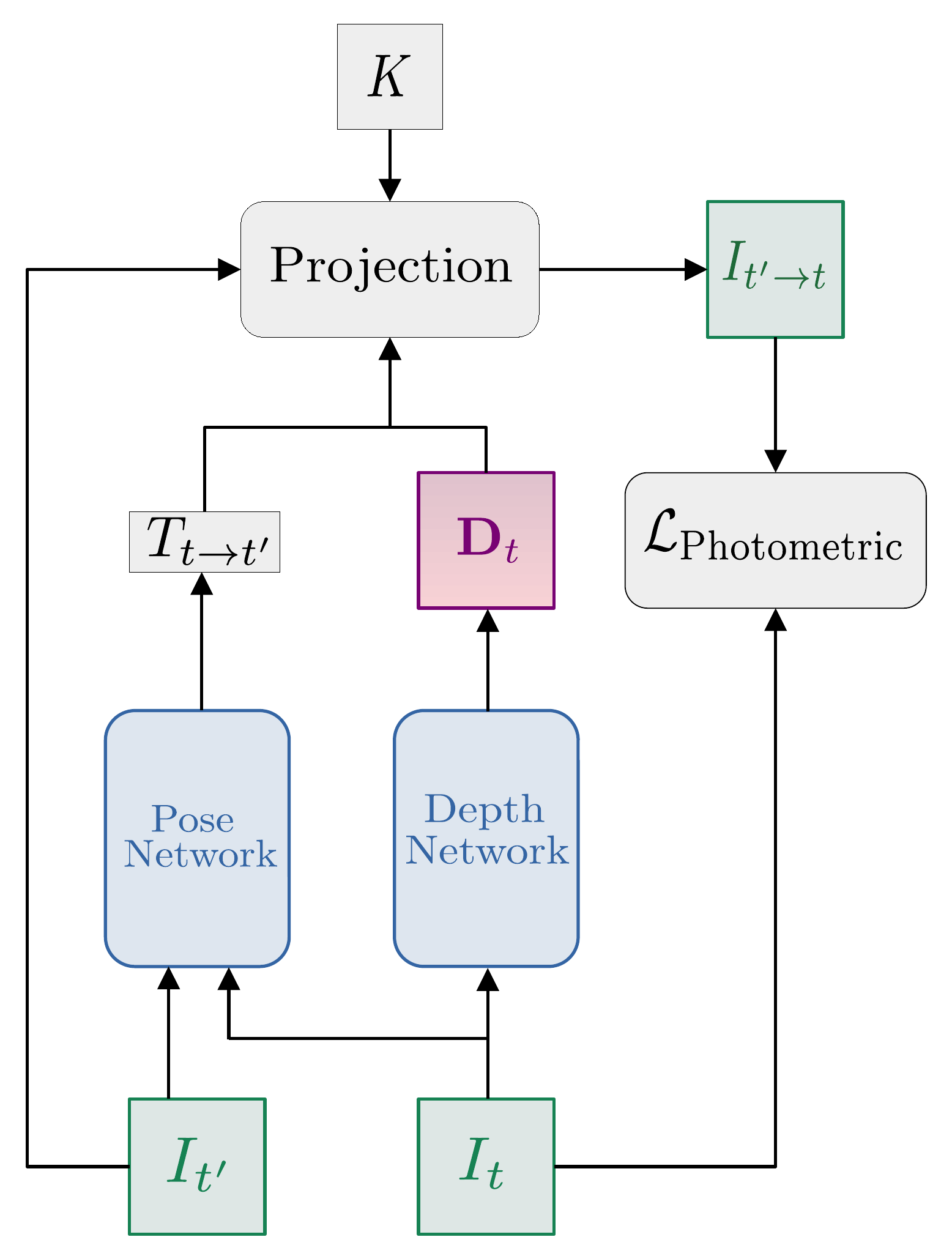}
\caption{The pose network computes the relative pose $T_{t \rightarrow t'}$ between the images $\rgbimage_t$ and $\rgbimage_{t'}$. The depth network computes the depth map $\depthmap_t$. Together with the intrinsic camera parameters $K$, we can use these estimates to warp $\rgbimage_{t'}$ onto $\rgbimage_t$ obtaining $\rgbimage_{t \rightarrow t'}$. The two images $\rgbimage_t$ and $\rgbimage_{t \rightarrow t'}$ are then compared by the means of a photometric loss (L1 loss and SSIM).}
\label{fig:photometric_loss}
\end{figure}

Despite certain specific differences, most of these training strategies are based on the same principle: 
\begin{itemize}
\item In a video sequence, we assume frames which are temporally close to each other to display the same scene.
\item We predict the depth map of one of the two frames, as well as their relative camera pose. 
\item We use these predictions and the intrinsic camera parameters to project one image onto the other. 
\item In order to train the depth prediction network and the pose prediction network, we compute a simple photometric loss between the warped and the target image.
\end{itemize}
This training strategy is visualized in Figure \ref{fig:photometric_loss} and more closely explained in the following text.  

Let $t$ and $t'$ be two close points in time in a video (e.g. at 3 frames distance). We assume that the corresponding frames $\rgbimage_t$ and $\rgbimage_{t'}$ display the same partially static scene, from two slightly different points of view. In order to warp the image $\rgbimage_{t'}$ onto $\rgbimage_t$, we will need the relative pose between the images. Since we only have single camera at our disposal (instead of two cameras with a fixed known distance), we will have to guess the relative camera position in the form of a rotation and a translation. This subtask is carried out by the means of a relative pose estimation network which outputs the transformation 
\begin{equation}
T_{t \to t'} = g(\rgbimage_{t}, \rgbimage_{t'})
\end{equation} between the two frames. More specifically, $T_{t \to t'}$ denotes the rotation and subsequent translation which are necessary to get from a point in $\rgbimage_t$ to the corresponding point in $\rgbimage_{t'}$. 

The second piece of information necessary for warping task will be the depth map 
\begin{equation}
\depthmap_t = h(\rgbimage_{t})
\end{equation}
which is predicted by the depth network $h$. As can be observed, the depth prediction only ever happens on single images, rendering the resulting trained network viable for single image input. Both networks, $g$ and $h$, will be trained jointly. 
The third necessary ingredient are the intrinsic camera parameters $K$, which are assumed to be known beforehand. 

We can then warp the image $\rgbimage_{t'}$ onto $\rgbimage_{t}$, obtaining
\begin{equation}
\rgbimage_{t' \to t} = \rgbimage_{t'} \left\langle \text{Projection}(\depthmap_t, T_{t \to t'}, K)  \right\rangle 
\end{equation}
where $\langle \cdot \rangle$ denotes a differentiable sampling operator and $\rgbimage_{t' \to t}$ is the result of warping the image $\rgbimage_{t'}$ onto $\rgbimage_t$. In order for the sampling operation to be differentiable (see \cite{spatial_transformer_nets, zhou2017unsupervised}), the pixel values are a linear interpolation of the four closest pixels at integer positions in the image from which we sample.  
The projection operation effectively allows us to compute for each coordinate $p_t$ in $\rgbimage_t$ the corresponding pixel position $p_{t \rightarrow t'}$ in $\rgbimage_{t'}$. This transformation of coordinates can be formulated (see \cite{zhou2017unsupervised}) as 
\begin{equation}
p_{t \rightarrow t'} \sim K T_{t \to t'} \depthmap_t K^{-1} p_t. 
\end{equation}

We now compare the two images $\rgbimage_{t \to t'}$ and $\rgbimage_t$ by the means of photometric loss, that expresses their difference. Minimizing this difference during training will imply the improvement of the two network-predicted components in this computation: the depth map and the relative pose. 
In our case of MonodepthV2 the image difference is expressed as     
\begin{equation}
\begin{split}
	\mathcal{L}_\text{Photometric} &= \frac{\alpha}{2}(1 - \text{SSIM}(\rgbimage_t, \rgbimage_{t' \rightarrow t})) \\
	&+ (1 - \alpha) ||\rgbimage_t - \rgbimage_{t' \rightarrow t}||_1, 
\end{split}
\end{equation}
where SSIM denotes the structural similarity index measure and $|| \cdot ||_1$ is a simple L1 loss. The authors of \cite{godard2019digging} set the relative loss weight $\alpha$ to $0.85$.
MonodepthV2 specifically uses an additional gradient based smoothing loss,  multiple neighbouring frames, and a masking scheme for pixel positions on which the feedback is deemed to be of insufficient quality. A detailed explanation of these auxiliary techniques would, however, go beyond the scope of conveying the general training idea. 

After training, the depth network is used in isolation for the purpose of obtaining pseudo depth maps on new images.

\end{document}